\documentclass[lettersize,journal]{IEEEtran}
\usepackage{amsmath,amsfonts}
\usepackage{algorithmic}
\usepackage{algorithm}
\usepackage{array}
\usepackage[caption=false,font=normalsize,labelfont=sf,textfont=sf]{subfig}
\usepackage{textcomp}
\usepackage{stfloats}
\usepackage{url}
\usepackage{verbatim}
\usepackage{graphicx}
\usepackage{cite}
\usepackage{multirow}
\usepackage[hidelinks]{hyperref}
\begin{document}


\title{DVLTA-VQA: Decoupled Vision-Language Modeling with Text-Guided Adaptation for Blind Video Quality Assessment}

\author{Li Yu, Situo Wang, Wei Zhou,~\IEEEmembership{Senior Member,~IEEE,} Moncef Gabbouj,~\IEEEmembership{Fellow,~IEEE,}
\thanks{This work was supported in part by the National Natural Science Foundation of China under Grant 62002172; and in part by The Startup Foundation for Introducing Talent of NUIST under Grant 2023r131.

Li Yu is with School of Computer Science, Nanjing University of Information Science \& Technology, Nanjing 210044, China, and also with Jiangsu Collaborative Innovation Center of Atmospheric Environment and Equipment Technology (CICAEET), Nanjing University of Information Science \& Technology, Nanjing, China (e-mail: li.yu@nuist.edu.cn).

Situo Wang is with School of Computer Science, Nanjing University of Information Science \& Technology, Nanjing 210044, China. (e-mail: 202212490261@nuist.edu.cn).

Wei Zhou is with the School of Computer Science and Informatics,
Cardiff University, CF244AG Cardiff, U.K. (e-mail:
zhouw26@cardiff.ac.uk)

Moncef Gabbouj is with the Faculty of Information Technology and Communication Sciences, Tampere University, Finland. (e-mail: moncef.gabbouj@tuni.fi).
}

}

\markboth{Journal of \LaTeX\ Class Files,~Vol.~14, No.~8, August~2021}%
{Shell \MakeLowercase{\textit{et al.}}: A Sample Article Using IEEEtran.cls for IEEE Journals}


\maketitle

\begin{abstract}

Inspired by the dual-stream theory of the human visual system (HVS)—where the ventral stream is responsible for object recognition and detail analysis, while the dorsal stream focuses on spatial relationships and motion perception—an increasing number of video quality assessment (VQA) works built upon this framework are proposed. Recent advancements in large multi-modal models, notably Contrastive Language-Image Pretraining (CLIP), have motivated researchers to incorporate CLIP into dual-stream-based VQA methods. This integration aims to harness the model's superior semantic understanding capabilities to replicate the object recognition and detail analysis in ventral stream, as well as spatial relationship analysis in dorsal stream. However, CLIP is originally designed for images and lacks the ability to capture temporal and motion information inherent in videos. 
To address the limitation, this paper propose a \underline{D}ecoupled \underline{V}ision-\underline{L}anguage Modeling with \underline{T}ext-Guided \underline{A}daptation for Blind Video Quality Assessment (DVLTA-VQA), which decouples CLIP's visual and textual components, and integrates them into different stages of the NR-VQA pipeline. Specifically, a Video-Based Temporal CLIP module is proposed to explicitly model temporal dynamics and enhance motion perception, aligning with the dorsal stream. Additionally, a Temporal Context Module is developed to refine inter-frame dependencies, further improving motion modeling. On the ventral stream side, a Basic Visual Feature Extraction Module is employed to strengthen detail analysis. Finally, a text-guided adaptive fusion strategy is proposed to enable dynamic weighting of features, facilitating more effective integration of spatial and temporal information.
Extensive experiments on multiple public VQA datasets demonstrate that the proposed method achieves state-of-the-art performance, significantly improving prediction accuracy and generalization capability. 
\end{abstract}


\section{Introduction}

\IEEEPARstart{W}{ith} the rapid growth of social networks and video sharing platforms, user-generated video content has increased exponentially~\cite{1}. This growth has led to significant variability in video quality, influenced by capture conditions and processing methods. Accurately assessing video quality is crucial not only for ensuring satisfactory viewing experiences, but also for optimizing video encoding, transmission, and processing algorithms~\cite{2,3}. 

Video quality assessment is typically categorized into three main types: full-reference video quality assessment (FR-VQA)~\cite{10,11,12,45}, reduced-reference video quality assessment (RR-VQA)~\cite{46,47,13}, and no-reference video quality assessment (NR-VQA)~\cite{17,18,9,20,21,22,31,32,34,35,36,37}. Traditional full-reference and reduced-reference methods rely on high-quality reference videos. However, such reference videos are often unavailable in real-world scenarios, particularly for the vast amount of video content captured by amateur users. As a result, no-reference video quality assessment (NR-VQA), which does not require reference videos, has become a focal point in research.

Early NR-VQA methods primarily rely on handcrafted features~\cite{17,18}, but lack generalizability. With the advent of deep learning and large-scale video datasets~\cite{38,39,40}, data-driven NR-VQA models~\cite{9,20,21,22,31,32,34,35,36,37} have demonstrated superior performance by automatically extracting quality-related features.
In recent years, human brain perception-inspired methods~\cite{7,10}, especially those based on the dual-stream theory of human vision, have made significant progress. The dual-stream theory suggests that the brain processes visual information through two distinct pathways: the dorsal stream (for spatial relationships and motion perception) and the ventral stream (for object recognition and detail analysis). This theoretical framework has inspired advances in computer vision, including video quality assessment. 

Based on this foundation, recent research has begun to integrate vision language models, such as CLIP~\cite{14}, into video quality assessment tasks to take advantage of their strengths in capturing high-level semantic features. By aligning visual and textual representations in a shared embedding space, CLIP facilitates a deeper understanding of both visual content and textual descriptions, allowing for a more comprehensive evaluation of video quality across multiple dimensions. CLIP has been initially introduced into image quality assessment (IQA)~\cite{4,5,44},  and later been extended to video quality assessment. Existing CLIP-assisted VQA methods primarily follow two strategies. The first strategy uses the CLIP-Visual module independently, relying solely on the visual encoder to extract video features. For instance, COVER~\cite{41} proposed a comprehensive video quality assessment framework, where CLIP-visual encoder extracts high-level semantic features, together with Swin Transformer and ConvNet processing other features. The second strategy uses the standard CLIP paradigm, combining its visual and textual encoders to calculate the similarity between visual features and textual prompts to assess video quality. For example, BVQI~\cite{6} integrates CLIP derived semantic metrics with spatio-temporal quality indicators, achieving competitive zero-shot performance without human annotations. Similarly, CLiF-VQA~\cite{7} introduces a CLIP-based semantic feature extractor (SFE), which slides over multiple regions of video frames to extract features associated with human perceptions.  Despite achieving notable results, both strategies face inherent limitations. Since CLIP-Visual is by-default designed for images and thus failing to capture inter-frame temporal relationships which are essential for videos. Meanwhile, current feature fusion strategies for above methods primarily rely on fixed paradigms, such as static weighting or simple concatenation. Such mechanisms lack adaptability to contextual semantics, which reduces their effectiveness for videos of complex contents.

To address these challenges, the \underline{D}ecoupled \underline{V}ision-\underline{L}anguage Modeling with \underline{T}ext-Guided \underline{A}daptation for Blind Video Quality Assessment (DVLTA-VQA) is proposed in this paper. The proposed approach decouples CLIP by utilizing its visual encoder for feature extraction and its textual encoder for feature fusion. In the feature extraction stage, the traditional CLIP-Visual module is enhanced by introducing the temporal analysis for videos, which captures coarse-grained temporal information across video frames. Additionally, the Temporal Context Module using Temporal Adaptive Convolution (TadaConv)\cite{28} is incorporated to refine inter-frame relationships at a finer granularity (analogous to the dorsal stream). Meanwhile, the Basic Visual Feature Extraction Module, focusing on detail analysis(analogous to the ventral stream), extracts detailed visual features specifically targeting spatial distortions. To unify the different perspectives above, a text-guided adaptive fusion strategy is used, using the CLIP textual encoder. This strategy dynamically adjusts feature integration based on semantic relevance, enhancing both the adaptability and perceptual alignment for the video quality prediction.  


The main contributions can be summarized as follows:

\begin{itemize}

\item[$\bullet$] {\bf{Decoupled CLIP-Based Framework:}} Inspired by the dual-stream theory, this paper proposes to decouple CLIP-visual and CLIP-textual to fit them into different stages of NR-VQA pipeline. The CLIP-Visual is used to extract high-level semantic features, while CLIP-Textual is used for dynamic and contextual guidance for feature fusion.

\item[$\bullet$] {\bf{Video-Based Temporal CLIP and Temporal Context Module:}} To enhance motion perception in the dorsal stream, this paper proposes the Video-Based Temporal CLIP, extending the image-oriented CLIP framework with temporal modeling capability. Additionally, the Temporal Context Module further refines fine-grained inter-frame contextual representations, enabling a more detailed and accurate exploration of motion dynamics. These two modules work synergistically to enhance motion perception for video quality assessment.

\item[$\bullet$] {\bf{Text-Guided Adaptive Fusion Strategy:}} By dynamically adjusting weights of features based on textual semantic information, the proposed text-guided adaptive fusion strategy derives a more precise and effective fusion results.

\end{itemize}

The remainder of this paper is organized as follows. Section II summarizes the related works of NR-VQA. Section III describes the proposed NR-VQA approach based on Video-Based Temporal CLIP and Text-Guided adaptive Fusion. Section IV presents the experimental results and detailed analysis. The paper is summarized in Section V.

\section{Related Work}
\subsection{NR-VQA Methods}
No-reference video quality assessment (NR-VQA) methods can be broadly categorized into traditional handcrafted approaches and deep learning-based methods. Traditional methods~\cite{17,18} rely on domain-specific knowledge to manually design features, such as spatiotemporal or distortion-specific metrics, which are then used for quality prediction. However, these methods require manual feature design, making them less adaptable to diverse distortion types, with limited generalization capabilities, particularly in complex video scenes. Consequently, these traditional methods have gradually been replaced by automated techniques, such as deep learning-based video quality assessment.

For deep learning-based NR-VQA methods, many researchers focus on extracting features from video frames, followed by temporal modules for quality assessment. For example, in~\cite{9}, video frame features are extracted via deep learning networks and combined with the Human Visual System’s (HVS) temporal perception mechanism. This approach integrates a GRU network, which handles long-term dependencies, with a subjectively motivated temporal pooling layer to address temporal lag effects. In~\cite{20}, a recurrent network is used to address the NR-VQA task, where video frames are input to a model for feature extraction, and the resulting feature vectors are then processed by an RNN-based temporal module to yield a final quality score. Similarly, 2BiVQA~\cite{36} employs a pre-trained CNN to extract patch-level features from video frames, while Bi-LSTM modules handle the spatial pooling of patches and temporal aggregation of frames, ultimately producing overall quality predictions via a fully connected layer. Given the exceptional performance of Transformers in handling time series data, researchers have started applying Transformers to video sequence quality assessments. In~\cite{21}, a Transformer is employed within the temporal module to integrate frame-level information for comprehensive video quality assessment.~\cite{22} presents an end-to-end NR-VQA model that utilizes 3D convolutional networks for feature extraction, followed by a Transformer for quality regression, culminating in a final quality score prediction via a fully connected layer. SSL-VQA~\cite{37} introduces semi-supervised learning with the Video Swin Transformer, capturing quality-aware representations through contrastive loss, while leveraging a self-supervised Spatio-Temporal Visual Quality Representation Learning (ST-VQRL) framework to enhance performance with limited human-annotated data.

Although deep learning methods have made significant advancements in NR-VQA, they predominantly rely on a single modality of visual features. As video content and scenes become increasingly complex, the limitations of using only visual features for quality assessment have become evident, especially when dealing with semantic information and complex scene understanding.

\begin{figure*}[t]
\centering
\includegraphics[width=7.2in]{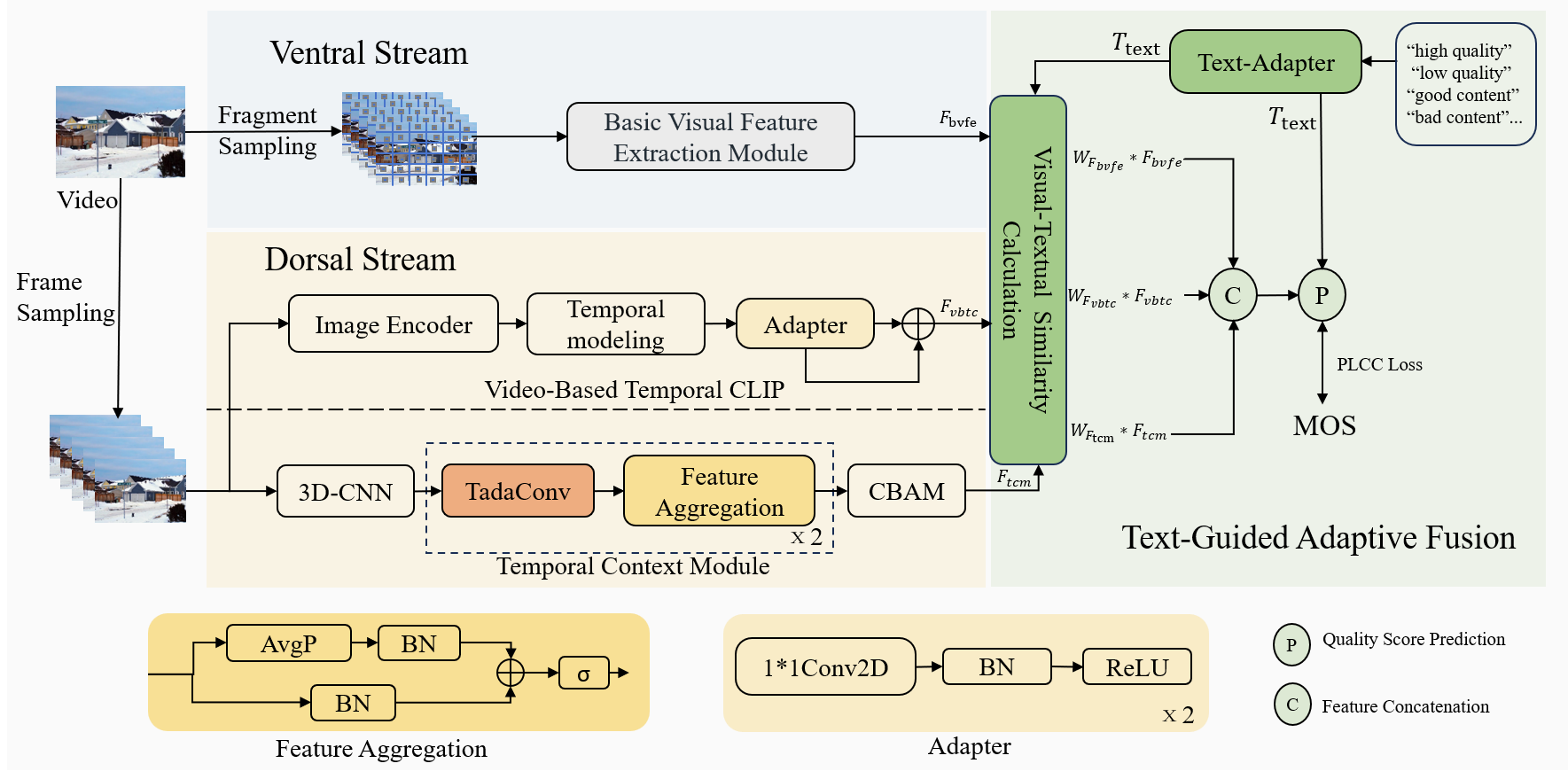}
\caption{The proposed method consists of Ventral Stream(blue), Dorsal Stream(yellow) and Text-Guided Adaptive Fusion(green). The Ventral Stream extracts low-level visual features through the Basic Visual Feature Extraction Module, while the  Dorsal Stream incorporates the Video-Based Temporal CLIP and Temporal Context Module for capturing high-level semantic features and fine-grained inter-frame temporal information. Features from both streams are then fused in the Text-Guided Adaptive Fusion block. The Text-Adapter integrates textual features to guide the fusion process, and the combined features are used for quality prediction.}
\label{frame}
\end{figure*}

\subsection{CLIP in NR-VQA}
Large vision-language models, particularly CLIP, have demonstrated exceptional cross-modal capabilities by effectively combining visual and textual features for semantic understanding.  These models have found widespread applications across numerous domains~\cite{49,50,51,52}.  In the field of quality assessment, CLIP-based methods were initially applied in image quality assessment (IQA) tasks. In these applications, customized prompts were utilized to evaluate and quantify image attributes such as appearance and content~\cite{4,5}. These approaches showcased CLIP’s ability to extract high-level semantic features, enabling it to outperform traditional methods in tasks requiring deep contextual understanding, much like the ventral stream’s role in object recognition.

In video quality assessment, CLIP has been adapted to encode frame-level information, treating video frames as independent image inputs. For instance,~\cite{6} combines CLIP-based semantic metrics with traditional spatiotemporal quality indicators, introducing a zero-shot video quality index (BVQI) that achieves competitive results without the need for annotated training data. Similarly,~\cite{7} develops a high-level semantic feature extractor (SFE) based on CLIP to align its perception with human quality judgments, leading to improved quality prediction accuracy. To enhance generalization performance,~\cite{23} integrates CLIP features with handcrafted or low-level features, such as those from FAST-VQA, achieving robust performance in assessing in-the-wild video content. However, existing CLIP-based VQA methods primarily process video frames independently, overlooking inter-frame temporal relationships, which limits their ability to model dynamic temporal changes, particularly motion perception, a function more aligned with the dorsal stream in HVS.

\subsection{Feature Fusion for NR-VQA}
Feature fusion is a critical component of multi-branch Video Quality Assessment models, which aim to capture a comprehensive representation of video quality by integrating spatial, temporal, and semantic information. For instance,~\cite{2} proposes a fusion strategy that combines spatial features extracted by Swin Transformer-B, motion features captured by SlowFast, and quality-aware features from Q-Align, LIQE, and FAST-VQA into a unified representation for predicting video quality. Similarly,~\cite{7} combines semantic and spatial features to create an overall representation, enabling effective quality prediction across different video datasets. Similarly,~\cite{24,25} utilizes a multi-branch structure to extract different features and fuses them into a final quality score. While these methods are effective and have shown promising results, their feature fusion strategies often rely on relatively simple approaches, such as concatenation or static weighting of features, or more advanced methods like feature fusion networks. These approaches struggle to adjust the importance of features dynamically according to their contextual relevance. This limitation underscores the need for more advanced fusion strategies that can adaptively prioritize features, focusing on the most critical aspects of video quality.

\section{Method}
Drawing inspiration from the dual-stream theory of the human visual system, our proposed framework aims to comprehensively analyze video quality by mimicking the ventral and dorsal streams, as illustrated in Fig. \ref{frame}. These streams specialize in detail analysis (ventral stream) and motion perception (dorsal stream), respectively. The input video is processed using two distinct sampling strategies: (1) Fragment Sampling: This involves cropping multiple small patches from video frames at their original resolution and concatenating them to form a fragment input. These fragments are then fed into the Basic Visual Feature Extraction Module for low-level feature extraction, mimicking the ventral stream's role in analyzing spatial details such as brightness, contrast, and texture. (2) Frame Sampling: This strategy selects a sequence of continuous video frames, which are passed through the Video-Based Temporal CLIP and the Temporal Context Module to capture high-level semantic and temporal features, thereby enhancing motion perception akin to the dorsal stream. Once features from both streams are extracted, they are integrated adaptively through the Text-Guided Adaptive Fusion module. In this stage, textual features from a text encoder are first processed by the Text Adapter. The cosine similarity between these textual features and the visual features from both streams is computed, enabling dynamic determination of fusion weights based on their relevance. Finally, the similarity between the fused video features and the text features is recalculated. This is followed by softmax pooling to produce the final quality score. The entire framework is trained using the PLCC loss function to optimize performance and ensure accurate quality assessment. This hierarchical approach effectively combines spatial detail analysis, motion perception, and semantic understanding to achieve a robust evaluation of video quality.

\subsection{Video-Based Temporal CLIP}

The image-based CLIP model has inherent limitations for video quality assessment tasks, primarily due to its lack of temporal modeling capabilities. While visual-language models designed for videos offer comprehensive video understanding, they typically come with an enormous number of parameters, making them unnecessarily complex for NR-VQA tasks. Instead, a lightweight model focusing on quality assessment-related features is needed. Therefore, this paper proposed the Video-Based Temporal CLIP module. This module inherits CLIP’s semantic feature extraction of image, and enhances its perception of video temporal dynamics through a lightweight temporal modeling mechanism and adapter, making it well-aligned with NR-VQA tasks.

Given an input video $V$, $T$ frames are first sampled to obtain an input video segment $V_{clip}=\{V_1, V_2, \dots, V _T\}$. These frames are processed independently by the visual encoder $f_\theta(\cdot)$ of the pre-trained CLIP model. For each frame, frame-level embeddings $z_i$ is generated, as follows:

\begin{equation}
z_i=f_\theta (V_i),
\end{equation}
where $i$ ranges from 1 to $T$. After generating the frame-level features $z_i$, temporal modeling is applied to aggregate them into a single global representation $Z$. This temporal modeling operation implicitly models temporal relationships, capturing coarse-grained temporal dynamics. Specifically, the temporal modeling operation is defined as:
\begin{equation}
Z=Pool(z_1,z_2,…,z_T),
\end{equation}
where $Pool(\cdot)$ refers to the average pooling operation.

To enhance the model's adaptability to the VQA task, an Adapter module is introduced after the temporal modeling layer. This adapter fine-tunes the global feature representation $Z$ using two convolutional blocks and combines the fine-tuned features with the original features through a residual connection. This allows the model to retain pre-trained knowledge while adapting to specific task-related features.

The adapter’s computation process is as follows. Given the feature $Z \in R^{(D,H,W)}$, the adapter is applied as:
\begin{equation}
A(Z)=Conv1(Conv2(Z^T)),
\end{equation}
where $Conv1$ and $Conv2$ are sequential combinations of 2D convolution, Batch Normalization (BN), and ReLU activation functions. Specifically, $Conv2$ reduces the channel dimension from \( D \) to \( \frac{D}{r} \), $Conv1$ restores the channel dimension back to \( D\). Employing this adapter enables the model to preserve robust feature representations.

To preserve the pre-trained knowledge from the CLIP model, a residual connection is used to combine the adapter’s output with the original feature $Z$. The residual connection is defined as:
\begin{equation}
F_{vbtc}=\alpha A(Z)+(1-\alpha)Z ,
\end{equation}
where $\alpha \in [0,1]$ is the weighting coefficient in the residual connection, controlling the fusion ratio between the fine-tuned features and the original features.

\subsection{Temporal Context Module}
While the Video-Based Temporal CLIP captures coarse temporal dynamics and video semantics, it lacks the ability to model fine-grained inter-frame temporal context. To address this, the Temporal Context Module is introduced, which employs 3D-CNNs and Temporal Adaptive Convolution (TadaConv)~\cite{28} to capture inter-frame temporal relationships with finer granularity, further enhancing motion perception akin to the dorsal stream.

To model fine-grained temporal context, a 3D-CNN is first used to process the input video clip $V_{clip} = \{V_1, V_2, \dots, V_T\}$, where each $V_i$ is a video frame. The 3D-CNN processes both the spatial and temporal dimensions of the frames simultaneously, generating a spatiotemporal feature representation $F_{ts}$. After obtaining the spatiotemporal features, TadaConv is applied to refine the temporal context. TadaConv dynamically adjusts the convolution kernel weights over time by introducing time-step calibration weights $\alpha _t$, thus enabling the model to adaptively capture complex temporal variations between frames. This process is formulated as follows:

\begin{equation}
\tilde{F_t} =W_t*F_{ts}=(\alpha_t \cdot W_b )*F_{ts} ,
\end{equation}
where $W_b$ represents the shared base convolution weights, and $\alpha_t$ is the dynamic time-step calibration weight, which adjusts the convolution operation at each time step to focus on the most relevant temporal features.

Next, the refined temporal features are aggregated using a feature aggregation module. To reduce redundancy and enhance feature expressiveness, temporal average pooling  is employed followed by normalization and activation operations: 
\begin{equation}
F_{agg} = \sigma (BN_1 (\tilde{F}_t)+BN_2 (TempAvgPool_p (\tilde{F}_t ))),
\end{equation}
where $BN_1$ and $BN_2$ are batch normalization operations, $TempAvgPool_p(\cdot)$ represents the temporal average pooling operation, and $\sigma(\cdot)$ is the ReLU activation function. This step ensures that important temporal features are preserved while reducing noise from redundant frame-level information.

To further enhance the temporal modeling capability, another layer of TadaConv is added for deeper temporal dynamic modeling. Following the dual processing by TAdaConv and feature aggregation, the features are passed to a Convolutional Block Attention Module (CBAM). CBAM enhances the model's focus on key spatiotemporal regions by combining channel and spatial attention mechanisms, thereby improving the accuracy of quality assessment. The detailed formula is as follows:
\begin{equation}
F_{tcm}=F_{agg}\cdot M_{channel} (F_{agg}) \cdot M_{spatial} (F_{agg}) ,
\end{equation}
where $M_{channel} (F_{agg})$ represents the channel attention weights, and $M_{spatial} (F_{agg})$  represents the spatial attention weights. The final feature $F_{tcm}$ obtained through this process serves as the output of the Temporal Context Module. 

\subsection{Basic Visual Feature Extraction Module}


To reinforce the ventral stream’s detail analysis, which focuses on spatial distortions and low-level visual features, FAST-VQA~\cite{27} is adopted as the Basic Visual Feature Extraction Module. FAST-VQA efficiently extracts quality-related low-level visual features, such as brightness, contrast, texture, and noise through local sampling and fragment processing. These features are indicative of spatial distortions and are crucial for assessing overall video quality.

 The feature extraction process begins with a preprocessing step, where fragment sampling is performed on the input video clip $V_{clip}$ to generate a series of local fragments $V_{fragments}$. Specifically, a unique fragment sampling strategy is used, where multiple small patches are cropped from video frames at the original resolution and then concatenated together. This strategy ensures that the distortions and quality characteristics of the video are preserved while allowing the model to process the data effectively.

These sampled fragments $V_{fragments}$ are then input into the FAST-VQA model, where the local visual features $F_{local}$ are extracted. The first step in this process involves applying a Swin Transformer to the local fragments, which is designed to capture local visual features through its hierarchical self-attention mechanism. This process not only preserves local information but also enables the model to capture features at multiple scales. Subsequently, the extracted local features $F_{local}$ are further processed by two 3D-CNN layers to generate the final global feature representation $F_{bvfe}$.The specific process is as follows:
\begin{equation}
F_{local}=Swin\textendash T(V_{fragments}) , 
\end{equation}
\begin{equation}
F_{bvfe}=Conv3D_2 (GELU(Conv3D_1 (F_{local}))) ,
\end{equation}
where $Conv3D_1$ and $Conv3D_2$ are two 3D convolution layers responsible for further refining local features along the temporal and spatial dimensions, and $GELU(\cdot)$ is the activation function. The final feature $F_{bvfe}$ , which captures low-level visual information from the video, will be integrated with features from other modules in the subsequent feature fusion process. 

\subsection{Text-Guided Adaptive Fusion}
In the realm of video quality assessment, conventional feature fusion methodologies predominantly employ rudimentary techniques such as feature concatenation or static weighting schemes. These approaches lack a dynamic adjustment mechanism that accounts for varying feature importance, which can diminish the precision of quality evaluations.
To tackle this challenge, we introduce an innovative text-guided adaptive fusion strategy. This method enables dynamic weight adjustments for visual features, thereby enhancing the interplay between visual and textual elements. By doing so, it significantly improves the model's capability to identify specific features associated with video quality, ultimately leading to more accurate assessments.

The weighting factors for the extracted temporal context features $F_{tcm}$, low-level visual features $F_{bvfe}$ , and high-level video semantic features $F_{vbtc}$ are determined by computing the cosine similarity between each visual feature and the textual feature. First, the text encoder of ViFi-CLIP~\cite{55} is used to extract the text embedding $T_{text}$, and the cosine similarity with each visual feature is calculated to obtain the weighting coefficients $W_{F_{bvfe}}$, $W_{F_{tcm}}$, and $W_{F_{vbtc}}$:

\begin{equation}
\begin{gathered}
W_{F_{bvfe}}=\frac{F_{bvfe}\cdot T_{\mathrm{text}}}{\Vert F_{bvfe}\Vert\Vert T_{\mathrm{text}}\Vert} \\
W_{F_{tcm}}=\frac{F_{tcm}\cdot T_{\mathrm{text}}}{\Vert F_{tcm}\Vert\Vert T_{\mathrm{text}}\Vert} \\
W_{F_{vbtc}}=\frac{F_{vbtc}\cdot T_{\mathrm{text}}}{\Vert F_{vbtc}\Vert\Vert T_{\mathrm{text}}\Vert}
\end{gathered}
\end{equation}
where $\cdot$ denotes the inner product, and $\Vert \cdot \Vert$ represents the L2 norm. By calculating the similarity between the visual features and the text embedding, a weighting factor is obtained for each visual feature, reflecting the importance of each feature within the current context.

Next, the calculated weighting factors are applied to the corresponding visual features, yielding the final fused feature:
\begin{equation}
F_{fused}=W_{F_{bvfe}} * F_{bvfe} + W_{F_{tcm}}*F_{tcm} + W_{F_{vbtc}} * F_{vbtc}
\end{equation}

\subsection{Quality Score Prediction}
To predict the final video quality score, the semantic alignment between the video features and textual descriptions of quality is leveraged. Positive and negative textual prompts related to video quality (e.g., “high quality” and “low quality”) are designed. The similarity between these textual prompts and the fused video features is then calculated. By applying softmax pooling to these similarities, the vision-language alignment is effectively transformed into a quality score.

Given the fused video feature representation $F_{fused}$, the positive and negative textual prompts are denoted as $P_a^+$ and $P_a^-$, respectively. The similarity $S_a^+$ and $S_a^-$ between the visual feature $F_{fused}$ and the positive and negative text features $T_{text}^+$ and $T_{text}^-$ are calculated as follows:
\begin{equation}
\begin{gathered}
S_a^+=Sim(T_{text}^+,F_{fused} ) \\
S_a^-=Sim(T_{text}^-,F_{fused})
\end{gathered} 
\end{equation}
where $Sim(\cdot)$ denotes the similarity calculation between text features and video features. Next, the softmax function is applied to these similarities to generate the final quality score $Q_{pre}$ as follows:
\begin{equation}
Q_{pre} = \frac{e^{S_a^+}}{(e^{S_a^+}+e^{S_a^-})}
\end{equation}

\subsection{Loss Function}
To maintain a high degree of linear correlation between the predicted scores and the subjective scores during model training, a loss function is constructed based on Pearson’s Linear Correlation Coefficient (PLCC). Given the predicted scores $Q_{\mathrm{pre}}$ and the ground-truth scores $Q_{\mathrm{gt}}$, the PLCC is defined as:
\begin{equation}\mathrm{PLCC}(Q_{\mathrm{pre}},Q_{\mathrm{gt}})=\frac{\mathrm{Cov}(Q_{\mathrm{pre}},Q_{\mathrm{gt}})}{\sqrt{\mathrm{Var}(Q_{\mathrm{pre}})}\sqrt{\mathrm{Var}(Q_{\mathrm{gt}})}}\end{equation}
where $\mathrm{Cov}(\cdot)$ and $\mathrm{Var}(\cdot)$ denote the covariance and variance, respectively. To enable the model to optimize video quality prediction by maximizing PLCC during backpropagation, it is incorporated into the loss function $\ell_{\mathrm{PLCC}}$, defined as follows:
\begin{equation}
\ell_{\mathrm{PLCC}} = 1 - \mathrm{PLCC}\bigl(Q_{\mathrm{pred}}, Q_{\mathrm{gt}}\bigr).
\label{eq:plcc_loss}
\end{equation}

By minimizing this loss, the model enhances the linear consistency between predicted and ground-truth scores during training, thereby enabling more accurate evaluation of different video quality levels.

\section{Experiments}
\subsection{Datasets}

To comprehensively evaluate the performance of the proposed DVLTA-VQA model, three essential VQA datasets are selected: KoNViD-1k~\cite{38}, LIVE-VQC~\cite{39} and YouTube-UGC~\cite{40}, as detailed in Table \ref{t0}. All of these datasets comprise user-generated content (UGC) videos featuring diverse scenes and various types of distortions.

\begin{table}[htp]
\caption{Details of Selected VQA Datasets}
\centering
\resizebox{0.47\textwidth}{!}{
\begin{tabular}{c|cccc}
\hline

Name & Total videos &   Resolution    &  Video Length(s)    & Distortion Type    \\
\hline

 KoNViD-1k  &    1200  &    540p     &     8  &    In-the-wild   \\

  LIVE-VQC  &    585  &    1080p-240p     &    10   &    In-the-wild    \\

  YouTube-UGC  &   1380   &  4k(HDR)-360p    &  20     &    In-the-wild    \\

\hline
\end{tabular}
}
\label{t0}
\end{table}

\noindent\textbf{KoNViD-1k~\cite{38}}. This dataset consists of 1,200 UGC videos collected from the YFCC100m~\cite{53} database, primarily in 720p resolution, with frame rates of 24fps, 25fps, or 30fps. The content covers a variety of natural scenes. Subjective quality scores were obtained through crowdsourcing, ensuring a wide range of evaluations and diversity. 

\noindent\textbf{LIVE-VQC~\cite{39}}. This dataset contains 585 videos shot by 80 users with 101 different devices across various scenes. The videos vary in resolution and shooting methods, and distortion types include overexposure, underexposure, and motion blur. The quality scores were also collected through crowdsourcing, with an average of 240 ratings per video.

\noindent\textbf{YouTube-UGC~\cite{40}}. This dataset includes 1,500 UGC videos selected from the YouTube platform, each 20 seconds in length, with resolutions ranging from 360p to 4K. Subjective quality scores were obtained from over 8,000 participants through crowdsourcing, ensuring data reliability.

\begin{table*}[htp]
\caption{Performance Comparison Results on Three Public Datasets, with the Best Results in Bold}
\centering
\begin{tabular}{c|cccccc}
\hline

Dataset & \multicolumn{2}{c}{KoNViD-1k}           & \multicolumn{2}{c}{LIVE-VQC}       & \multicolumn{2}{c}{YouTube-UGC}       \\

            Methods               & SROCC            & PLCC            & SROCC            & PLCC            & SROCC            & PLCC            \\

\hline
TLVQM~\cite{17}                    & 0.7730          & 0.7680          & 0.7990          & 0.8030          & 0.6690          & 0.6590          \\
GSTVQA~\cite{35}                    & 0.8140          & 0.8250          & 0.7880          & 0.7960          & N/A          & N/A          \\
VIDEVAL~\cite{18}                    & 0.7830          & 0.7800          & 0.7520          & 0.7510          & 0.7790          & 0.7730          \\
FASTVQA~\cite{27}                    & 0.8900          & 0.8890          & 0.8540          & 0.8520          & 0.8570          & 0.8530          \\
FasterVQA~\cite{29}                  & 0.8950          & 0.8980          & 0.8430          & 0.8580          & 0.8630          & 0.8590          \\
SimpleVQA~\cite{34}                  & 0.8560          & 0.8600          & N/A             & N/A             & 0.8470          & 0.8560          \\
PVQ~\cite{30}                  & 0.7910          & 0.7860          & 0.8270             & 0.8370             & N/A          & N/A          \\
VSFA~\cite{9}                  & 0.7730          & 0.7750          & 0.7730             & 0.7950             & 0.7240          & 0.7430          \\
CoINVQ~\cite{24}                  & 0.7670          & 0.7640          & N/A             & N/A             & 0.8160          & 0.8020          \\
DisCoVQA~\cite{31}                  & 0.8460          & 0.8490          & 0.8200             & 0.8260             & 0.8090          & 0.8080          \\
BVQA~\cite{32}                       & 0.8330          & 0.8340          & 0.8400          & 0.8500          & 0.8160          & 0.8040          \\
MaxVQA~\cite{23}                     & 0.8940          & 0.8950          & 0.8540          & 0.8730          & 0.8940          & 0.8900          \\
CLiFVQA~\cite{7}                    & 0.9020          & 0.9030          & 0.8640          & 0.8800          & 0.8880          & 0.8910          \\
DOVER~\cite{33}                      & 0.8970          & 0.8990          & 0.8120          & 0.8520          & 0.8790          & 0.8830 \\
COVER~\cite{41}                      & 0.8933          & 0.8947          & 0.8093          & 0.8478          & 0.9143          & \textbf{0.9165} \\
\hline
\textbf{DVLTA-VQA (Ours)}              & \textbf{0.9121} & \textbf{0.9127} & \textbf{0.8918} & \textbf{0.9027} & \textbf{0.9180} & 0.9053    \\

\hline
\end{tabular}

\label{t1}
\end{table*}

\subsection{Evaluation Metrics}
Two commonly used objective performance metrics are employed to assess the model: the Spearman Rank Order Correlation Coefficient (SROCC) and the Pearson Linear Correlation Coefficient (PLCC). These metrics evaluate the model's predictive performance from different perspectives, defined as follows:

\noindent\textbf{SROCC (Spearman Rank Order Correlation Coefficient)}. This metric measures the rank consistency between predicted scores and ground truth scores, evaluating the monotonic relationship between two sequences. SROCC values range from [-1, 1], where values closer to 1 indicate greater alignment between predicted and actual rankings, reflecting the model’s ability to preserve ranking consistency.

\noindent\textbf{PLCC (Pearson Linear Correlation Coefficient)}. PLCC evaluates the linear correlation between predicted and actual scores, reflecting the precision of the model's predictions. PLCC values also range from [-1, 1], with values closer to 1 indicating higher prediction accuracy.

\subsection{Implementation Details}

The experiments were conducted on a server equipped with an NVIDIA RTX 4090 GPU, implemented using the PyTorch framework. For the semantic feature extraction module, the pre-trained ViFi-CLIP model was loaded, and its weights were frozen to extract high-level semantic information from the videos. Similarly, the Basic Visual Feature Extraction Module used the pre-trained FAST-VQA model with frozen weights. The training process employed the AdamW~\cite{48} optimizer with an initial learning rate set to 1e-3. The weight coefficient $\alpha$ in the residual connection was set to 0.4.

\subsection{Performance Comparison}

In the overall performance comparison experiments, the proposed DVLTA-VQA is evaluated against several state-of-the-art video quality assessment approaches. These include vision-language-based methods such as BVQA~\cite{32}, MaxVQA~\cite{23}, CLiF-VQA~\cite{7}, and COVER~\cite{41}, as well as traditional deep learning approaches without vision-language models, such as FAST-VQA~\cite{27}, FasterVQA~\cite{29}, and SimpleVQA~\cite{34}. The experiments were conducted on three widely-used benchmark datasets: KoNViD-1k, LIVE-VQC, and YouTube-UGC, with performance measured in terms of SROCC and PLCC.

As shown in Table \ref{t1}, the proposed DVLTA-VQA outperforms other approaches in both SROCC and PLCC across all three datasets, demonstrating higher prediction accuracy and ranking consistency. On the KoNViD-1k dataset, the proposed method surpasses the second-best approach by 1.1\%. On the LIVE-VQC dataset, it achieves the highest performance, outperforming CLiF-VQA by 3.2\% in SROCC and 2.6\% in PLCC. On the YouTube-UGC dataset, the proposed method also shows strong competitive performance, further validating its capability to handle challenging user-generated content. These results validate the effectiveness of the proposed dual-stream architecture in capturing both ventral stream-inspired spatial details and dorsal stream-inspired temporal dynamics, leading to significantly improved video quality prediction and validating the proposed feature fusion strategy.

\begin{figure}[htp]
\centering
\subfloat[]{\includegraphics[width=1.7in]{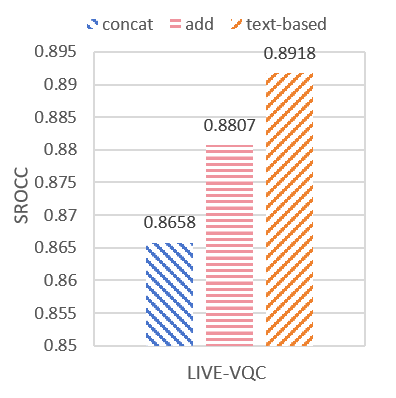}
\label{fig:subfig1ab}}
\subfloat[]{\includegraphics[width=1.7in]{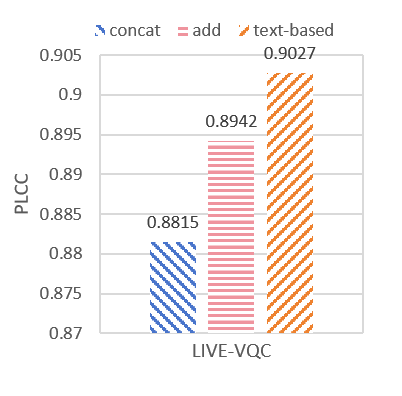}
\label{fig:subfig2ab}}

\caption{Ablation results of different feature fusion strategies on the LIVE-VQC dataset.} 
\label{fab}
\end{figure}

\begin{table}[htp]
\caption{Ablation Results on Temporal Feature Extraction Module}
\centering
\begin{tabular}{c|cc}
\hline

\multirow{2}{*}{\textit{}} & \multicolumn{2}{c}{LIVE-VQC}                 \\
                           & SROCC            & PLCC                   \\

\hline

c3d                 & 0.8895         & 0.8943               \\
r(2+1)d             & 0.8848     & 0.8959         \\
tadaconv        & \textbf{0.8918} & \textbf{0.9027}   \\

\hline
\end{tabular}

\label{t3}
\end{table}

\begin{table}[htp]
\caption{Ablation Results on Feature Extraction Branches (BVFEM: Basic Visual Feature Extraction Module, TCM: Temporal Context Module, VTC: Video-Based Temporal CLIP)}
\centering
\begin{tabular}{ccc|ccccccc}
\hline
   \multirow{2}{*}{BVFEM
}   &\multirow{2}{*}{TCM}  & \multirow{2}{*}{VTC}  & \multicolumn{2}{c}{LIVE-VQC}            \\ 

  &    &  &  SROCC  & PLCC \\ \hline

      \checkmark     &  \checkmark   &        & 0.8614  &   0.8660    \\

        \checkmark  &     &     \checkmark   & 0.8440   &  0.8520  \\

          &  \checkmark   &    \checkmark    &  0.8800  &   0.8853   \\

       \checkmark   &  \checkmark   &    \checkmark    &  \textbf{0.8918} &   \textbf{0.9027}   \\

     \hline
\end{tabular}
\label{t4}
\end{table}

\subsection{Ablation Study}

\noindent\textbf{Ablation Study on Feature Fusion}. To verify the effectiveness of the proposed text-guided adaptive fusion strategy, an ablation study was conducted comparing it with two commonly used fusion methods: direct concatenation (concat) and simple weighted addition (add). As shown in Fig. \ref{fab}, on the LIVE-VQC dataset, the text-guided adaptive fusion strategy achieved the best performance in both SROCC and PLCC metrics. These results indicate that the text-guided adaptive fusion strategy can more flexibly capture the importance of each feature, better meeting the needs of video quality assessment. Compared to traditional methods, the proposed fusion strategy more effectively integrates video semantic features, temporal dynamic features, and low-level visual features, significantly improving model performance.

\noindent\textbf{Ablation Study on Temporal Context Module}. In the Temporal Context Module, the TAdaConv convolution layer adaptively models inter-frame temporal information to capture finer-grained dynamic changes. In this ablation study, TAdaConv was replaced with standard 3D convolutional structures such as C3D and R(2+1)D to analyze the relative advantages of TAdaConv in temporal dynamic modeling. As shown in Table \ref{t3}, the use of TAdaConv achieved the best results, confirming its unique advantage in handling temporal dynamics. By adaptively adjusting convolution kernel weights, TAdaConv more effectively captures dynamic features between video frames, thereby improving the accuracy of video quality assessment.

\noindent\textbf{Ablation Study on Different Modules}. To validate the contribution of each module in the proposed dual-stream architecture, a comprehensive ablation study was conducted by systematically removing one module at a time from the full model while keeping the other two modules intact. The results are presented in Table \ref{t4}, which compares the performance of the full model with configurations where one module is removed. This design provides a more comprehensive understanding of each module's impact on the video quality assessment task. 
Notably, the removal of the Temporal Context Module caused the most significant performance drop, emphasizing its critical role in capturing fine-grained temporal dynamics essential for accurate video quality assessment. In contrast, the complete three-module model achieved the highest SROCC and PLCC scores of 0.8918 and 0.9027, respectively, showcasing the superiority of integrating temporal, semantic, and low-level visual features. These findings validate the necessity of all three modules, as their combined contributions deliver the most accurate and robust performance in video quality assessment, aligning with the proposed framework’s design inspired by the dual-stream theory of human vision.

\begin{figure*}[htp]
\centering
\subfloat[]{\includegraphics[height=1.5in, width=2.4in]{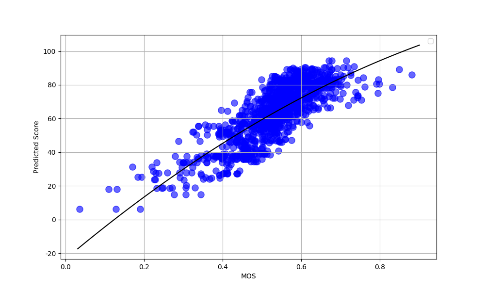}
\label{fig:subfig1m}}
\subfloat[]{\includegraphics[height=1.5in, width=2.4in]{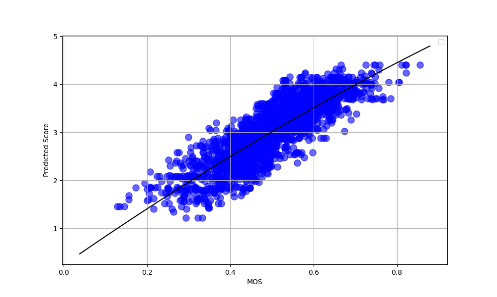}
\label{fig:subfig2m}}
\subfloat[]{\includegraphics[height=1.5in, width=2.4in]{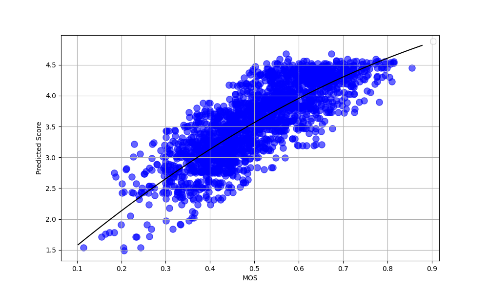}
\label{fig:subfig3m}}

\caption{Scatter plots of subjective quality scores versus predicted scores across different datasets. (a)LIVEVQC. (b) KoNVid-1k. (c) YouTube-UGC.}
\label{f4}
\end{figure*}

\begin{table*}[ht]
\caption{Generalization evaluations among existing in-the-wild VQA databases}
\centering
\begin{tabular}{c|cccccccccccc}
\hline
Training Set  & \multicolumn{4}{c}{KoNViD-1k}                                  & \multicolumn{4}{c}{LIVE-VQC}                                          & \multicolumn{4}{c}{Youtube-UGC}                                   \\
Test Set      & \multicolumn{2}{c}{LIVE-VQC} & \multicolumn{2}{c}{Youtube-UGC} & \multicolumn{2}{c}{KoNViD-1k}     & \multicolumn{2}{c}{Youtube-UGC}   & \multicolumn{2}{c}{LIVE-VQC}      & \multicolumn{2}{c}{KoNViD-1k} \\
\hline
    Method          & SROCC          & PLCC         & SROCC           & PLCC           & SROCC            & PLCC            & SROCC            & PLCC            & SROCC            & PLCC            & SROCC          & PLCC          \\
\hline

TLVQM         & 0.5730        & 0.6290       & 0.3540         & 0.3780         & 0.6400          & 0.6300          & 0.2180          & 0.2500          & 0.4880          & 0.5460          & 0.5560        & 0.5780        \\
CNN-TLVQM     & 0.7130        & 0.7520       & 0.4240         & 0.4690         & 0.6420          & 0.6310          & 0.3290          & 0.3670          & 0.5510          & 0.5780          & 0.5880        & 0.6190        \\
VIDEVAL       & 0.6270        & 0.6540       & 0.3700         & 0.3900         & 0.6250          & 0.6210          & 0.3020          & 0.3180          & 0.5420          & 0.5530          & 0.6100        & 0.6200        \\
MDTVSFA       & 0.7160        & 0.7590       & 0.4080         & 0.4430         & 0.7060          & 0.7110          & 0.3550          & 0.3880          & 0.5820          & 0.6030          & 0.6490        & 0.6460        \\
GST-VQA       & 0.7000        & 0.7330       & N/A             & N/A             & 0.7090          & 0.7070          & N/A              & N/A              & N/A              & N/A              & N/A            & N/A            \\
BVQA          & 0.6950        & 0.7120       & 0.7800         & 0.7800         & 0.7380          & 0.7210          & 0.6020          & 0.6020          & 0.6890          & 0.7270          & 0.7850        & 0.7820        \\
DisCoVQA      & 0.7820        & 0.7970       & 0.4150         & 0.4490         & 0.7920          & 0.7850          & 0.4090          & 0.4320          & 0.6610          & 0.6850          & 0.6860        & 0.6970        \\
MaxVQA        & 0.7930        & 0.8320       & 0.8670         & 0.8570         & 0.8330          & 0.8310          & 0.8460          & 0.8240          & 0.8040          & 0.8120          & 0.8550        & 0.8520        \\
\textbf{DVLTA-VQA} & \textbf{0.8134}  & \textbf{0.8404} & \textbf{0.8678}   & \textbf{0.8593}   & \textbf{0.8527} & \textbf{0.8459} & \textbf{0.8488} & \textbf{0.8460} & \textbf{0.8062} & \textbf{0.8207} & \textbf{0.8579}  & \textbf{0.8561} \\
\hline
\end{tabular}
\label{t5}
\end{table*}

\begin{figure*}[t]
\centering
\includegraphics[height=3.3in,width=6.8in]{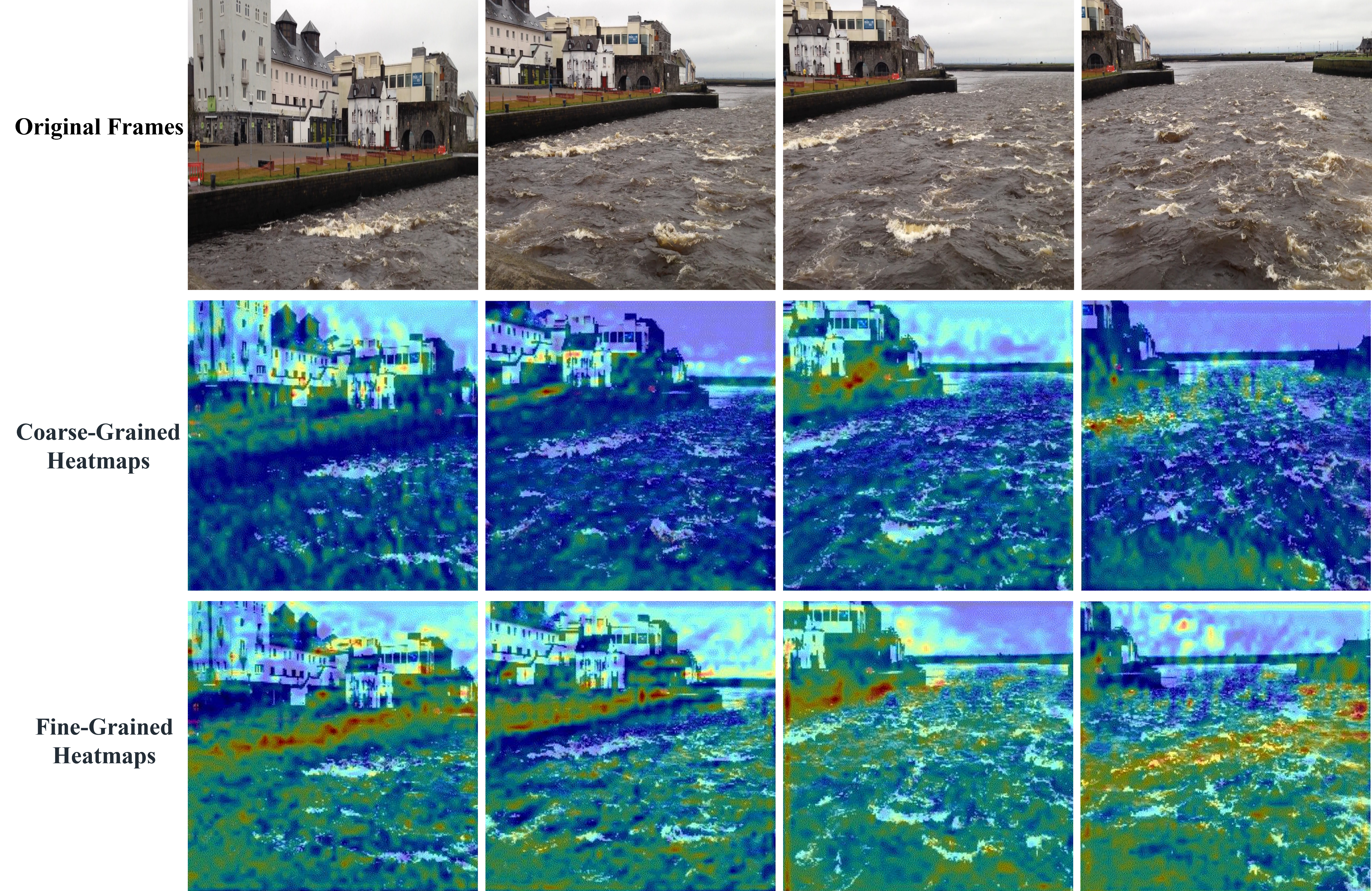}
\caption{This figure illustrates the differences in attention between coarse-grained and fine-grained temporal dynamic information. The top row shows original video frames. The second row shows heatmaps highlighting the model's focus on prominent motion, such as the moving shoreline, while the bottom row demonstrates the model's ability to capture finer details, including the continuously moving waves.}
\label{v_f}
\end{figure*}

\subsection{Visualization}
To provide a clearer illustration of the relationship between the predicted results and subjective scores, a visualization was conducted, as shown in Fig. \ref{f4}. The figure presents scatter plots for the LIVE-VQC, KoNVid-1k, and YouTube-UGC datasets, which intuitively reflect the consistency between the predicted scores and subjective ratings. To further clarify this relationship, logistic fitting curves are used to depict the trend between the predicted scores and subjective ratings. The scatter plots show that most of the predicted scores are closely aligned with the fitting curves, indicating a high level of agreement between the model’s predictions and subjective scores across different datasets, thus validating the accuracy and effectiveness of the proposed method.

\subsection{Heatmap Visualization}
To compare the differences between coarse-grained and fine-grained temporal dynamic information, heatmap visualization of the model’s features was performed, as shown in Fig. \ref{v_f}. The top row displays the original video frames. The second row shows heatmaps corresponding to coarse-grained features, while the third row visualizes fine-grained features. Coarse-grained features are obtained through temporal pooling, which aggregates frame-level features into a single representation. This enables the model to focus on prominent motions or global semantic features. As shown in the visualization (second row), the model primarily attends to the moving shoreline, while paying limited attention to the surrounding waves.

In contrast, fine-grained features are captured through temporal convolution mechanisms, allowing the model to dynamically adapt to small-scale temporal variations, which complements the motion perception ability of the dorsal stream. The visualization (bottom row) highlights this capability, demonstrating the model's detailed attention not only to the shoreline's motion but also to the continuously moving waves, reflecting its ability to capture fine-grained temporal dynamic information.

\subsection{Cross Database Performance}
An ideal video quality assessment model should exhibit excellent generalization capability, meaning it can maintain stable performance when evaluating unseen and untrained data samples. To verify whether the proposed method possesses this key attribute, a cross-dataset testing experiment was designed and conducted on three datasets: KoNViD-1k, LIVE-VQC, and YouTube-UGC. Specifically, the model was first trained on one video quality assessment dataset, and the trained model was then tested on two other distinct video datasets. This experimental setup provides a clear evaluation of the model’s adaptability and stability across different datasets.

As shown in Table  \ref{t5}, the experimental results reveal that the proposed NR-VQA method consistently outperforms existing approaches on the two test datasets. These results not only confirm the strong generalization capability of the model but also demonstrate the significant advantages of integrating semantic understanding and spatiotemporal feature processing in video quality assessment tasks. Notably, the proposed Video-based CLIP module and Temporal Context Module play a critical role in extracting high-level semantic features, capturing inter-frame dynamics, and enhancing contextual representation between frames. Furthermore, the text-guided adaptive fusion strategy effectively optimizes the feature integration process, significantly improving the accuracy and robustness of video quality evaluation.

\section{Conclusions}
This paper proposes a novel decoupled-CLIP-based NR-VQA model, DVLTA-VQA, inspired by the dual-stream theory of HVS. The proposed method effectively addresses challenges in inter-frame dynamics modeling and feature fusion faced by traditional approaches. By decoupling the CLIP visual and textual components and innovatively introducing Video-Based Temporal CLIP and Temporal Context Module, the framework significantly enhances the model’s ability to perceive temporal information in videos. Combined with a text-guided adaptive fusion strategy, it achieves more efficient feature integration. Experimental results fully demonstrate the effectiveness and superiority of the proposed framework in video quality assessment tasks.
This research not only provides a new and effective method for no-reference video quality assessment but also offers strong evidence for the application of human visual mechanisms to video quality assessment. Exploring further brain-inspired models and biological mechanisms may lead to new breakthroughs in video quality assessment and offer valuable insights for broader applications in video processing and cross-modal learning.

\bibliographystyle{IEEEtran}
\bibliography{paper}

\vfill

\end{document}